\pdfoutput=1

\documentclass[11pt]{article}

\usepackage[]{acl}

\usepackage{times}
\usepackage{latexsym}
\usepackage{enumitem}

\usepackage[T1]{fontenc}

\usepackage[utf8]{inputenc}

\usepackage{microtype}

\usepackage{inconsolata}

\newcommand{\projectname}{\mbox{Zero-NatVer}}

\usepackage{float}
\usepackage{multicol}
\usepackage{microtype}
\usepackage{graphicx}
\usepackage{booktabs}
\usepackage{amssymb}
\usepackage{amsmath}
\usepackage{multirow}
\usepackage{booktabs}
\usepackage{comment}
\usepackage{tikz}
\usepackage{listings}
\usepackage{xcolor}
\usepackage{array}
\usepackage{makecell}
\usepackage{lipsum}
\usepackage{cuted}
\usepackage{dirtytalk}

\usepackage{siunitx}
\newcommand{\tbf}{\textbf}

\usepackage{easyReview}

\usepackage{soul}

\definecolor{listingbg}{rgb}{0.97,0.97,0.97}
\definecolor{listingexample}{rgb}{0.55,0.7,0.9}
\definecolor{codegreen}{rgb}{0,0.6,0}
\definecolor{codegray}{rgb}{0.5,0.5,0.5}
\definecolor{codepurple}{rgb}{0.58,0,0.82}
\lstdefinestyle{mystyle}{
    backgroundcolor=\color{listingbg},
    commentstyle=\color{codegreen},
    keywordstyle=\color{magenta},
    numberstyle=\tiny\color{codegray},
    stringstyle=\color{codepurple},
    basicstyle=\fontsize{8pt}{9pt}\sffamily,
    breakatwhitespace=false,
    breaklines=true,
    captionpos=b,
    keepspaces=true,
    showspaces=false,
    showstringspaces=false,
    showtabs=false,
    tabsize=2,
    breakindent=0pt,
    frame=lines,
    aboveskip=\baselineskip,
    framesep=10pt,
    captionpos=b,
    abovecaptionskip=10pt,
    moredelim=**[is][\color{listingexample}]{@}{@},
    columns=fullflexible,
    numbers=none,
    xleftmargin=2pt,
    framexleftmargin=8pt,
}
\title{Zero-Shot Fact Verification via Natural Logic and Large Language Models}

\author{Marek Strong, Rami Aly, Andreas Vlachos \\
    Department of Computer Science and Technology \\
    University of Cambridge \\
    \{ms2518,rmya2,av308\}@cam.ac.uk}

\begin{document}
\maketitle

\begin{abstract}

The recent development of fact verification systems with natural logic has enhanced their explainability by aligning claims with evidence through set-theoretic operators, providing faithful justifications.
Despite these advancements, such systems often rely on a large amount of training data annotated with natural logic. To address this issue, we propose a zero-shot method that utilizes the generalization capabilities of instruction-tuned large language models.
To comprehensively assess the zero-shot capabilities of our method and other fact verification systems, we evaluate all models on both artificial and real-world claims, including multilingual datasets.
We also compare our method against other fact verification systems in two setups.
First, in the \textit{zero-shot generalization} setup, we demonstrate that our approach outperforms other systems that were not specifically trained on natural logic data, achieving an average accuracy improvement of \num{8.96} points over the best-performing baseline.
Second, in the \textit{zero-shot transfer} setup, we show that current systems trained on natural logic data do not generalize well to other domains, and our method outperforms these systems across all datasets with real-world claims.

\end{abstract}

\section{Introduction}


In the context of fact-checking, fact verification (FV) is a process of verifying whether a textual hypothesis holds, based on retrieved evidence.
While many improvements have been made in this field due to the recent rapid growth in NLP \citep{akhtar-etal-2023-multimodal,guo2022survey,nakov2021automated}, FV systems often employ pipelines with black-box components that hide the underlying reasoning. 

One line of research attempts to improve explainability with attention-based methods \citep{shu2019defend,popat2018declare} and post-hoc summarizations \citep{atanasova2020generating,kotonya2020explainable_a}. However, these approaches do not provide \textit{faithful justifications} --- explanations that accurately reflect the model's decision-making process and the data it used \citep{jacovi2020towards}.
In contrast, systems such as NaturalLI \citep{angeli2014naturalli} and ProoFVer \citep{krishna2022proofver} provide faithful justifications by expressing semantic relations between claim/evidence pairs. Modeling these logical relations and their aggregation explicitly with natural logic (NatLog) allows for the accurate processing of phenomena such as double-negation and has resulted in more accurate and robust fact-checking systems.

However, a limitation of natural logic-based FV systems is that they require large amounts of training data annotated with entire natural logic proofs. For example, ProoFVer \citep{krishna2022proofver} was trained on 145K instances artificially obtained from structured knowledge bases such as PPDB \citep{ganitkevitch2013ppdb} and Wikidata \citep{vrandevcic2014wikidata}. While recent work \citep{aly2023qa} attempts to alleviate this issue by proposing a few-shot learning method trained on as few as \num{32} instances, human annotation of even a small number of proofs can be impractical and expensive, as it requires substantial linguistic knowledge and familiarity with natural logic.
Moreover, few-shot systems 
require additional training data in order to generalize effectively to new domains, further increasing the costs.

\begin{figure*}[th!]
\centering
\includegraphics[width=1.0\linewidth, trim={0.5cm 0.1cm 0.5cm 0.1cm},clip]{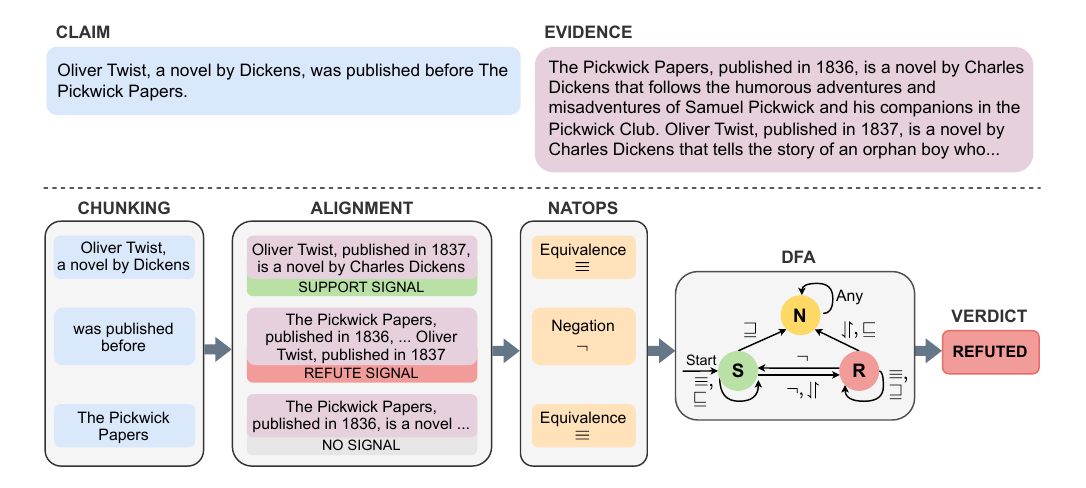}
\caption{\textbf{Proof generation with natural logic in \projectname{}.} Initially, the claim and evidence texts are chunked and aligned. \projectname{} then assigns natural logic operators (NatOps), using a QA framework and alignment signals parsed from the previous step. This process produces a proof sequence comprising \textit{(claim, evidence, NatOp)} triples. Lastly, NatOps act as transitions in the DFA, with the final state (here~Refuted) determining the verdict.}
\label{fig:highlevel}
\end{figure*}

To this end, we propose \textbf{\projectname{}}\footnote{Code is available at: \url{https://github.com/marekstrong/Zero-NatVer}}, a zero-shot fact verification approach for constructing natural logic proofs that leverages prompting and question-answering with instruction-tuned large language models (LLMs). 
\projectname{}'s proof generation process is illustrated in Figure \ref{fig:highlevel}. First, a claim is chunked into smaller units of information. Then, the units are aligned to relevant parts of the evidence, and natural-logic operators are assigned to each claim-evidence pair. Lastly, the proofs are executed on a finite state automaton (DFA) as defined in natural logic inference, producing the verdict.

Unlike previous NatLog-based approaches, our method also addresses the problem of limited context during the NatOp assignment stage by producing alignment signals (e.g., support and refute) and passing them to the next stage for NatOp assignments. This enables more accurate NatOp predictions. Additionally, \projectname{} uses constrained decoding to prevent hallucinations, and it uses question-answering (QA) ensembles to reduce the variability of predictions.

We evaluate our method on real-world and artificial FV datasets, including Climate-FEVER \cite{diggelmann2020climate}, PubHealth \cite{kotonya2020explainable_a}, SciFact \cite{wadden2020fact}, and Hover \cite{jiang2020hover}. We also demonstrate that \projectname{} can generalize to \mbox{non-English} datasets by evaluating the system on the Danish dataset DanFever \citep{norregaard2021danfever}, Mandarin Chinese dataset CHEF (\citep{hu2022chef}), Arabic dataset Unified-FC \cite{baly2018integrating}, and the Russian/Ukrainian portion of the dataset RU22Fact \cite{zeng2024ru22fact}.
In a zero-shot setup, where models have not been trained on any data labeled with natural logic, our approach outperforms all NatLog baselines by \num{8.96} accuracy points when averaged across all English datasets. It is also competitive with the direct QA approach, where the model is prompted directly for an answer.
Thus, our method, which is based on natural logic, provides both improved performance on unseen domains and explainability via faithful justifications.

\section{Related Work}
\label{sec:related_work}

Natural logic \citep{VanBenthem1986, sanchez1991studies} and NaturalLI \citep{angeli2014naturalli}, composes full inference proofs that operate directly on natural language, capable of expressing more complex logical relationships between claim and evidence, such as double-negation. \citet{krishna2022proofver} trained natural logic inference systems for fact verification, achieving competitive performance while remaining faithful and more explainable than its entirely neural counterpart. While these neural-symbolic approaches require substantial training data to perform well, \citet{aly2023qa} explored natural logic inference in a few-shot setting by casting natural logic operators into a question-answering framework, subsequently making use of the generalization capabilities of instruction-tuned language models. 
Although our work also considers question-answering, we further expand on this approach, addressing prediction calibration issues frequently encountered in a zero-shot setting \citep{kadavath2022language, jiang2023calibrating}.
Other neuro-symbolic reasoning systems for FV use simple logical rules to aggregate veracity information on a claim's components to provide simple faithful explanations \citep{stacey-etal-2022-logical, stacey2023logical, Chen_Bao_Sun_Zhang_Chen_Zhou_Xiao_Li_2022}. However, these rules lack the expressiveness of natural logic and thus cannot inherently model more complex phenomena such as double negation.



Previous work on zero-shot FV is limited and largely relies on the generation of weakly supervised training samples and on knowledge of the target domain \citep{pan-etal-2021-zero, wright-etal-2022-generating}. \citet{pan2023investigating} observe that typical FV systems fail when transferred to unseen domains in a zero-shot setting and propose a data augmentation technique to improve generalizability.
Moreover, none of the aforementioned zero-shot methods produces (faithful) explanations.
In a few-shot setting, several recent works have explored the use of large language models that produce explanations alongside the verdict. \citet{pan-etal-2023-fact} define a reasoning program consisting of a sequence of subtasks to verify complex claims. \citet{yao2023react} propose chain-of-thought prompting complemented by action operations to support the model's reasoning and its explanation generation. \citet{li2023chain} propose to edit rationales generated via chain-of-thought prompting by querying knowledge sources.
Unlike our work, these approaches still rely on in-context examples. 

\section{Zero-NatVer}

Given a claim $c$ and evidence sentences $e_{1}, e_{2}, ..., e_{k} \in E$, our system determines the veracity label $y$, which denotes whether the information from $E$ supports $c$, refutes $c$, or whether there is not enough information to reach a verdict. \projectname{} obtains the verdict in four steps, executed by an instruction-tuned LLM. 

In the first two steps, \projectname{} segments $c$ into several chunks (Sec.~\ref{sec:chunking}) and aligns each such chunk with relevant information from $E$ (Sec.~\ref{sec:alignment}).
This process results in a sequence of $l$ claim-evidence alignment pairs $A=a_{1}, a_{2}, ..., a_{l}$.
As part of this alignment process, we also generate alignment explanations that are parsed for supporting/refuting signals. These signals are used in the third stage of the pipeline where \projectname{} determines semantic relations of aligned pairs in terms of natural logic. Thus, it generates a sequence of natural logic operators $O=o_{1}, o_{2}, ..., o_{l}$, which correspond to alignment pairs in $A$ (Sec.~\ref{sec:method_ensembles}).
Finally, $O$ is used in the last stage to traverse a deterministic finite automaton (DFA), which determines the claim's veracity. The following sections describe each step in more detail.

\subsection{Chunking}
\label{sec:chunking}

\begin{figure}[!ht]
\centering
\includegraphics[width=1.0\linewidth, trim={0.7cm 0.5cm 0.7cm 0.5cm}, clip]{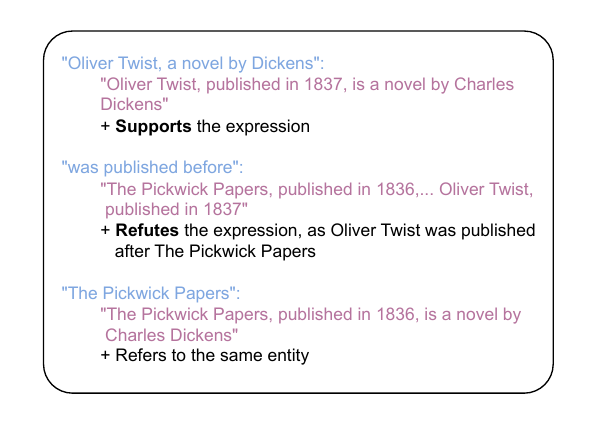}
\caption{\textbf{Claim-evidence alignments with explanations.} The blue text indicates provided claim chunks. The purple text represents generated evidence alignments, and the black text denotes alignment explanations, which are later parsed for signals.}
\label{fig:chalign_alignment}
\end{figure}

FV systems that are based on natural logic split claims into smaller, more manageable pieces, also called chunks \citep{krishna2022proofver}. These chunks, typically consisting of only a few words, represent a single atomic piece of information that can be independently verified and linked to relevant information in the evidence text.

We perform this task by prompting an LLM to \textit{"Split the claim text into smaller chunks that can be individually fact-checked."} We then use constrained decoding to ensure the desired output format. Specifically, the model is allowed to either generate consecutive characters from the provided text or insert a special token (e.g., a newline character) to denote the start of a new chunk. This process is executed as follows:
\begin{enumerate}[noitemsep, topsep=0pt, partopsep=0pt]
    \item The claim text $c$ is pre-processed as a queue of tokens $Q_{C}$.
    \item The decoding is prefixed with an initial phrase to encourage the generation of claim chunks.
    \item The model is constrained to sample only one of two outputs - the next token from $Q_{C}$ or a newline character.
    \item Repeats step 3 until $Q_{C}$ is empty (i.e., all claim tokens are consumed).
\end{enumerate}

Given the constraints at each decoding step, the model cannot hallucinate new words, skip words, or alter information in the claim.

\subsection{Alignment}
\label{sec:alignment}

In the second stage of the pipeline, each previously generated claim chunk is aligned with the corresponding information in the provided evidence sentences. We use an LLM to perform this alignment by prompting it with $c$, $E$, and all claim chunks (see details in Appendix \ref{sec:appendix-prompting}).
Furthermore, we prompt the model to also generate alignment explanations for each generated alignment. Figure \ref{fig:chalign_alignment} shows an example of the model's output.

To enforce the expected output format, we use constrained decoding, switching between three decoding modes: \textit{claim}, \textit{evidence}, and \textit{\mbox{alignment-explanation}}. In the claim mode, we simply insert the chunk text, and no further text is generated. In the evidence mode, the model generates the alignment and is constrained so that it cannot use tokens that occur only in $C$ and not in $E$. This constraint is meant to reduce hallucinations and prevent the model from aligning chunks with claim tokens. Lastly, the inference process is not constrained in the alignment-explanation mode because explanations are only searched for keywords and are not used in the following stages or as part of the proof.

Although constraint decoding helps mitigate hallucinations, it is important to note that the model could still hallucinate in evidence mode, as it is allowed to generate words not present in either $C$ or $E$. Indeed, we analysed all alignments and found out that \num{12.4}\% of chunks contained at least one token absent from $E$. To solve this issue, we post-process the alignments and remove all text that does not form sequences of tokens in evidence sentences $E$. This post-processing step ensures that the alignment process is faithful and that only information from the evidence is used to verify the claim. Alternatively, we could constrain the decoding process to generate only tokens present in the evidence text. However, our empirical findings showed that this approach struggles in situations where it needs to combine two or more pieces of information that are not adjacent in the evidence text.

Lastly, the alignment explanations are parsed for supporting and refuting signals, which are used by the NatOp assigner. A simple keyword search was sufficient to effectively determine the signals while prioritizing precision over recall.

\begin{figure*}[ht!]
\centering
\includegraphics[width=0.94\linewidth, trim={0.5cm 0.4cm 0.5cm 0.3cm},clip]{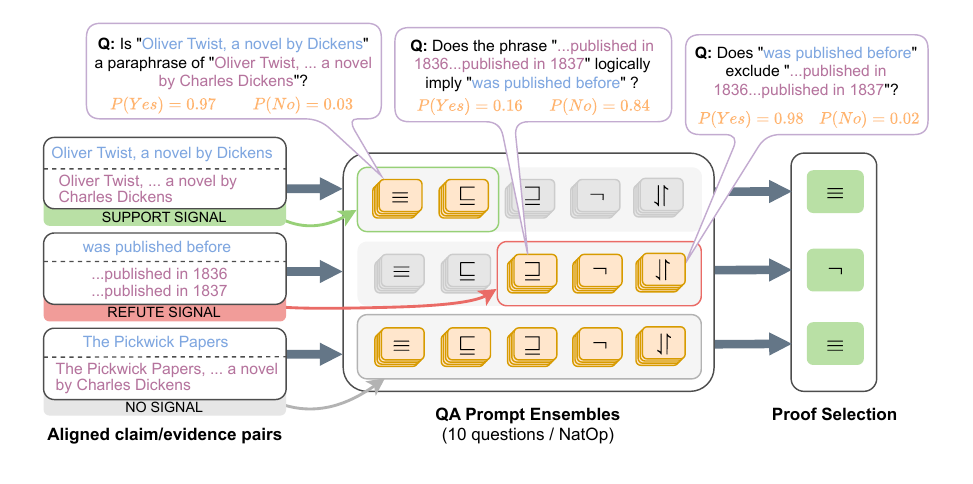}
\caption{\textbf{Proof generation process of \projectname{}.} First, we utilize alignment signals, where available, to identify the set of potential NatOp candidates (represented by orange blocks). Next, we apply prompt ensembles and NatOp priority to select the final NatOp (depicted as green blocks).}
\label{fig:proof_pipeline}
\end{figure*}

\subsection{NatOp Assignment via QA Ensembles}
\label{sec:method_ensembles}

\begin{table}[b]
        \resizebox{1.0\linewidth}{!}{
        \begin{tabular}{ >{\centering\arraybackslash}m{24mm}| >{\centering\arraybackslash}m{20mm}|m{40mm} } 
            \hline
            \textbf{NatOp} & \textbf{Definition} & \textbf{Template Example} \\ 
            \hline
            \hspace{0.5em}Equivalence\newline($\equiv$)
            & $x=y$
            & Is X a paraphrase of Y? \\
            \hline
            Forward Entailment\newline($\sqsubseteq$)
            & $x \subset y$
            & Given the premise X does the hypothesis Y hold?\\
            \hline
            Reverse Entailment\newline($\sqsupseteq) $
            & $x \supset y$
            & Does the expression Y entail X?\\
            \hline
            \hspace{0.75em}Negation\newline($\neg$)
            & $x\cap y=\varnothing \wedge x\cup y = U$
            & Is the phrase X a negation of Y?\\
            \hline
            \hspace{0.75em}Alternation\newline($\downharpoonleft \! \upharpoonright$)
            & $x\cap y=\varnothing \wedge x\cup y\neq U$
            & Does X exclude Y?\\
            \bottomrule
        \end{tabular}
        }
    \caption{Natural logic operators (NatOps) with set-theoretic definitions and template examples.}
    \label{tab:natop-examples}
\end{table}


Once the claim and evidence are aligned, the next step is to determine a single NatOp for each claim-evidence pair, which represents the semantic relation between the corresponding chunks. 

We start by preparing the list of NatOp candidates for each alignment pair, considering five basic operators, as shown in Table \ref{tab:natop-examples}.
This process is guided by alignment signals from the previous stage, and we define the candidate lists as follows:

\begin{itemize}[noitemsep, topsep=2pt, partopsep=2pt]
    \item For a supporting signal, we use operators that indicate the evidence chunk entails the claim chunk: $[\equiv, \sqsubseteq]$.

    \item For a negative signal, we use operators that indicate the claim chunk is not entailed by the information in the evidence chunk: $[\neg, \sqsupseteq, \downharpoonleft \! \upharpoonright]$.

    \item In case of no signal, the full set of NatOps is used: $[\equiv, \neg,  \sqsubseteq, \sqsupseteq, \downharpoonleft \! \upharpoonright]$.
\end{itemize}

This process allows for transferring some global information from the aligner, which has access to the full claim and evidence texts, to the NatOp assigner, which only sees chunks and thus has limited knowledge. For example, in Figure \ref{fig:highlevel}, the aligner aligns \textit{"was published before"} with corresponding years for each publication, describing the ordering of events. While this alignment is reasonable for a reader with access to the entire claim and evidence texts, it becomes challenging to determine its meaning if we only see the aligned sub-strings.

For each aligned pair, we then consider operators in the corresponding candidate lists, and this process is detailed in Figure \ref{fig:proof_pipeline}. Similar to \citet{aly2023qa}, we treat these operators as relations that can be inferred via questions over claim-evidence spans.
Thus, we prompt our model with \textit{Yes/No} questions to determine whether a relation can be expressed by one of the NatOps. 
If none of these operators is successfully determined by the QA framework, we assign the independence \mbox{operator $\#$}, which implies that there is no semantic relation.


In order to reduce the variability of outcomes, we use a large number of \textit{Yes/No} questions to prompt the model, thereby obtaining several micro-judgements per NatOp, which are then aggregated as a weighted average.
In our experiments, we employ \num{10} templates for each NatOp. Rather than manually hand-crafting these question templates, we employ the LLM to generate them. Consequently, this approach allows for easy generation of additional templates as needed.



For a given claim-evidence alignment pair $a$ and operator $o$, we compute a NatOp score $s_{o,a}$ as a weighted average over all micro-judgments:
\begin{align}
    \label{eq:natop_score}
    s_{o,a} = \sum_{i=1}^{N} w_{i}\text{ QA}(\text{Yes}|T_{i},a)
\end{align}
where $T$ is a collection of prompt templates, and $w$ represents confidence weights for each template, with $\sum_{i=1}^N w_i = 1$.

We compute $w_i$ by iterating over the entire dataset in a single pass and capturing the log-likelihood scores for each template. For each instance, we always capture only the Yes/No option, which has the higher log-likelihood score (i.e.,\ the option that the model favors more).

Using Equation \ref{eq:natop_score}, we then compile a list of NatOps candidates $C$, considering only those where $s_{o,a} > \alpha$, with $\alpha$ serving as a confidence threshold. Since we are not using any validation data to determine hyperparameters, we set $\alpha = 0.5$, as we are considering only two output classes.

Due to the ambiguity of natural language and the complexity of alignments, it frequently occurs that $|C|>1$. Therefore, we must resolve this conflict and select a single NatOp from $C$. However, we want to minimize the likelihood of incorrectly choosing NatOps that lead to the \textit{Not Enough Evidence} state, from which there are no outgoing transitions to other states. 
Thus, we use a NatOp priority approach, selecting from the operators in the following order: $[\equiv, \neg, \sqsubseteq, \sqsupseteq, \downharpoonleft \! \upharpoonright]$. We defined this order by considering the natural ordering of relations described in \citet{icard2012inclusion}.
For instance, in a scenario where the candidate list $C$ consists of equivalence ($\equiv$) and alternation ($\downharpoonleft \! \upharpoonright$), we postulate that identifying equivalence (i.e., assessing textual similarity) is a simpler task compared to identifying alternation (i.e., recognizing non-exhaustive exclusion). This order was determined prior to our experiments and was not further optimized.

\section{Experimental Methodology}

\subsection{Zero-Shot Setups}

To better assess the zero-shot capabilities of our approach, we differentiate between two types of zero-shot setups-- \textbf{zero-shot generalization} and \textbf{zero-shot transfer}. We define zero-shot generalization as a model's ability to handle entirely new tasks or domains it has not encountered during training. Conversely, zero-shot transfer refers to training a model on a specific task or dataset and subsequently applying it to a different but related task or dataset without further training. For example, consider a model trained on a broad spectrum of general data (e.g., BART, T5, or Llama) that did not include proofs with natural logic. Applying this model to FV with natural logic then exemplifies zero-shot generalization according to our definition.
In contrast, if the same model is fine-tuned on a dataset annotated with natural logic proofs and then applied to perform FV with natural logic on a different dataset, this would be an instance of zero-shot transfer.






\begin{table*}
\centering
\resizebox{0.85\linewidth}{!}{
    \setlength\tabcolsep{5pt}
    \renewcommand{\arraystretch}{1.1}
    \begin{tabular}{l |c ||cc |cc |cc |cc }
    \hline
        \multirow{3}{*}{\textbf{System}} & \multirow{3}{*}{\textbf{Model}} & \multicolumn{2}{c|}{\textbf{C-FEVER}} & \multicolumn{2}{c|}{\textbf{SciFact}}  & \multicolumn{2}{c|}{\textbf{PubHealth}}  & \multicolumn{2}{c}{\textbf{Hover}}\\
        & &  F1 & Acc & F1 & Acc & F1 & Acc & F1 & Acc    \\
    \hline
    ProoFVer    & BART& 26.63 & 34.75          & 25.58 & 34.67     & 38.15 & 39.27      & 47.13 & 49.76 \\
    QA-NatVer   &  Flan-T5 & 22.20 & 36.86    & 23.56 & 40.67     & 44.42 & 48.73      & 35.65 & 50.85 \\
    QA-NatVer   & Llama3-8B & 32.6 & 36.5    & 37.18 & 43.67     & 63.66 & 68.79      & 49.95 & 54.93 \\
    \tbf{\projectname{}}  & Llama3-8B   & 46.02& 51.12& \tbf{54.58}& \tbf{58.33}& 69.21& 70.01& \tbf{60.26}& \tbf{60.27}\\
    \hline
    Direct-QA & Llama3-8B & \tbf{51.27}  & \tbf{58.58}  & 52.76 & 57.00 &  \tbf{78.18} & \tbf{78.18} &  55.34 & 57.00 \\
    \hline
    \hline
    Full Supervision & - & 75.7  & -  & 71.1 & - & 85.88 & 86.93 & - & 81.2 \\
    \hline
    \end{tabular}
}
\caption{\textbf{Zero-shot generalization results for English datasets.} Macro-F1 and accuracy scores for systems that were \textbf{not} specifically trained on FV datasets. Where possible, we also report available SOTA results with fully-supervised models trained on in-domain data as a reference.}
\label{tab:main_table_results_generalisation}
\end{table*}

\begin{table*}
\centering
\resizebox{0.85\linewidth}{!}{
    \noindent
    \setlength\tabcolsep{4pt}
    \renewcommand{\arraystretch}{1.1}
    \begin{tabular}{l |c |c || cc |cc |cc |cc}
    \hline
        \multirow{2}{*}{\textbf{System}} & \multirow{2}{*}{\textbf{Model}} & \multirow{2}{*}{\textbf{\makecell[bc]{Train size\\(FEVER)}}}& \multicolumn{2}{c|}{\textbf{C-FEVER}}  &  \multicolumn{2}{c|}{\textbf{SciFact}}&\multicolumn{2}{c|}{\textbf{PubHealth}}   & \multicolumn{2}{c}{\textbf{Hover}}\\
        &  && F1 & Acc &  F1 & Acc &F1 & Acc & F1 & Acc     \\
    \hline
    Pan et al.& BERT  & 800& 40.60   & -      &  50.71     & -  &60.06 & -    & - & - \\
    ProoFVer            & BART&145K& 40.70  & 43.35     &  45.57  & 49.16   &57.78 & 61.22   & 57.08  & 57.89   \\
    QA-NatVer         & Flan-T5 &64& 44.74  & 47.43      &  52.02  & 56.67   & 61.8 & 61.8   & \tbf{62.44}  & \tbf{63.48} \\
    \tbf{\projectname{}}  & Llama3-8B    &None& \tbf{46.02}& \tbf{51.12}&  \tbf{54.58}& \tbf{58.33}& \tbf{69.21}& \tbf{70.01}& 60.26& 60.27 \\
    \hline
    \end{tabular}
}
\caption{\textbf{Zero-shot transfer results for English datasets.} Macro-F1 and accuracy scores for systems trained on the FEVER dataset. For each system, we report the provided language model and the size of the training data. The results presented in Pan et al. (2013) do not include accuracy scores and do not cover the Hover dataset. }
\label{tab:main_table_results_transfer}
\end{table*}

\subsection{Datasets}

Previous studies on NLI-based FV models have primarily focused on evaluating performance using artificial claims from FEVER-like datasets \cite{krishna2022proofver,aly2023qa,chen2023converge}. However, these datasets typically encompass only general topics, and artificial claims tend to be structurally simple. To achieve a more comprehensive assessment of zero-shot capabilities, we have evaluated our models on both artificial and natural claims, including non-English datasets.


For artificial claims, we evaluated models using claims from the multi-hop dataset Hover \citep{jiang2020hover} and the Danish dataset \mbox{DanFEVER} \cite{norregaard2021danfever}. For real-world claims, we included English datasets \mbox{Climate-FEVER} \citep{diggelmann2020climate}, PubHealth \citep{kotonya2020explainable_a}, and SciFact \citep{wadden2020fact}, as well as the non-English datasets CHEF \citep{hu2022chef}, Unified-FC \cite{baly2018integrating}, and RU22Fact \cite{zeng2024ru22fact}. For datasets that provide knowledge bases for retrieval, we used BM25 \citep{robertson1994some} to retrieve evidence. Further details are provided in Appendix \ref{sec:appendix_data}.


\subsection{Baselines}

Our NatLog baselines consist of ProoFVer \citep{krishna2022proofver} and QA-NatVer \citep{aly2023qa}. We always aim to use the largest possible backbone LLMs to make our results more comparable. However, both baseline models have specific limitations due to their current implementations.

ProoFVer currently supports only models from the Fairseq1 toolkit\footnote{https://github.com/facebookresearch/fairseq}, and the largest supported model is BART \citep{lewis2019bart}. For zero-shot transfer setups, we use ProoFVer with BART, which was trained on 145K FEVER instances. For non-English datasets, we use mBART \citep{liu2020multilingual} instead.

QA-NatVer can use larger LLMs, such as Flan-T5 \citep{chung2022scaling}, but its implementation currently supports training only for encoder-decoder model architectures. Therefore, we were unable to fine-tune QA-NatVer with Llama3 for zero-shot transfer experiments and instead used Flan-T5 trained on \num{64} instances. For experiments on DanFEVER, we used the mT0 \citep{muennighoff2022crosslingual} backbone. The zero-shot generalization setup does not require any training, so we were able to use Llama3-8B for inference.

For a non-NatLog baseline, we use the Llama3-8B model, prompting it to directly assign a verdict (i.e., \textit{Supported}, \textit{Refuted}, or \textit{Not Enough Information}), based on the provided claim and evidence texts. We refer to this baseline as Direct-QA. The prompting details are described in Listing \ref{prompt_direct_mc}.

Additionally, we include results reported by \citet{pan2023investigating} as a further baseline for zero-shot transfer experiments. More details about our baselines can be found in Appendix \ref{sec:appendix_baselines}.

\subsection{Implementation Details}

We conducted our main experiments with the Llama3-8B model \citep{llama3modelcard, dubey2024llama}. Crucially, we did not fine-tune the model on any specific dataset, and we did not tune any hyperparameters. The only exposure to fact-checking datasets was when we were designing our prompts. For this purpose, we used a separate dataset, Symmetric-Fever \citep{schuster2019towards}. We selected a small subset of \num{100} claims and tested that our prompts generated responses in the desired format. For hyperparameters, we have adopted the recommendations of \citet{perez2021true} and did not rely on hyperparameters from prior works.
Further details are provided in Appendix \ref{sec:appendix-models}.

\section{Results}

\subsection{Zero-Shot Generalization}

\label{sec:main_results}

We report the main results for zero-shot generalization in Table ~\ref{tab:main_table_results_generalisation}. \projectname{} consistently outperforms other NatLog-based baselines across all datasets, including both synthetic and real-world claims. Averaging results across all datasets, it achieves an accuracy of \num{59.93} points, surpassing ProoFVer by \num{20.32} accuracy points. When compared to the version of QA-NatVer that uses the same backbone model (Llama3-8B) as Zero-NatVer, our method demonstrates an average improvement of \num{8.96} accuracy points.



We also report results for the Direct-QA setup, a non-NatLog approach, where the Llama3-8B model directly assigns the verdict. Table~\ref{tab:main_table_results_generalisation} shows that \projectname{} outperforms Direct-QA on SciFact and Hover, demonstrating its competitive performance while improving the model's explainability through the generation of proofs. Additionally, the results for Direct-QA might be overly optimistic, given that Llama3 was trained on 15 trillion tokens, making it likely that some datasets were included in its training data.
Since \projectname{} does not use Llama3 to directly predict verdicts, and the final verdict is derived from NatLog proofs, its performance is likely to be more representative.


We also report state-of-the-art (SOTA) results for each dataset to highlight the performance gap between models fully supervised on in-domain data and zero-shot approaches. Each reported SOTA result comes from a separately trained model, and there is no guarantee that this performance will generalize to other datasets or languages. The reported metrics, including F1 and Accuracy scores where available, represent the best results to our knowledge. Our findings indicate that \projectname{} helps close this gap while maintaining the advantage of using a single model that does not require fine-tuning.

\subsection{Zero-Shot Transfer}

We report the main results for zero-shot transfer in Table ~\ref{tab:main_table_results_transfer}.
\projectname{} consistently outperforms both ProoFVer and the results reported by \citet{pan2023investigating} across all datasets, despite these baselines being trained on NatLog data and ProoFVer's substantial training set of \num{145}K instances. These findings highlight the robust generalization capabilities of Llama3, which our method effectively leverages.

\projectname{} also surpasses QA-NatVer on all datasets except Hover, exceeding QA-NatVer by an average of \num{2.59} accuracy points. This indicates that while NatLog baselines trained on FEVER data generalize effectively to similar domains, such as Hover and DanFEVER (the latter is discussed further below), their performance does not extend well to real-world claims. Therefore, in practical applications, it may be more advantageous to allocate computational resources to more powerful language models rather than to fine-tuning.



\begin{table*}
\centering
\resizebox{0.85\linewidth}{!}{
    \setlength\tabcolsep{5pt}
    \renewcommand{\arraystretch}{1.1}
    \begin{tabular}{l |c ||cc |cc |cc |cc}
    \hline
        \multirow{3}{*}{\textbf{System}} & \multirow{3}{*}{\textbf{Model}} & \multicolumn{2}{c|}{\textbf{DanFEVER}} & \multicolumn{2}{c|}{\textbf{CHEF}} &\multicolumn{2}{c|}{\textbf{Unified-FC}} &\multicolumn{2}{c}{\textbf{RU22Fact}}\\
 & & \multicolumn{2}{c|}{Da} & \multicolumn{2}{c|}{Zh} & \multicolumn{2}{c|}{Ar} & \multicolumn{2}{c}{Ru/Ukr}\\
        & &  F1 & Acc    & F1 & Acc    & F1 & Acc    & F1 & Acc    \\
    \hline
    ProoFVer    & mBART& 29.8& 41.97&20.16&38.57 &49.18&49.85 &43.66&57.74\\
    QA-NatVer   & mT0 & 35.68& 37.05&-&- &-&- &-&-\\
    QA-NatVer   & Llama3-8B & 48.92& 55.35&-&- &-&- &-&-\\
    \tbf{\projectname{}}  & Llama3-8B   & \tbf{53.9}& \tbf{62.55}& \tbf{47.94}& \tbf{53.2} &57.23&57.35 &79.89&86.57\\
    \hline
    Direct-QA & Llama3-8B & 52.77 & 61.7 & 19.5 & 24.04  &\tbf{62.42}&\tbf{63.98} &\tbf{84.41}&\tbf{87.95}\\
    \hline
    \hline
    Full Supervision & - & 90.2& -&67.62&- &89.9&91.0 &60.56&-\\
    \hline
    \end{tabular}
}
\caption{\textbf{Zero-shot generalization results across non-English multilingual datasets.} Macro-F1 and accuracy scores for systems that were \textbf{not} specifically trained on FV datasets. QA-NatVer currently does not support non-English languages, except for Danish. Where possible, we also report available SOTA results with fully-supervised models trained on in-domain data as a reference.}
\label{tab:multilingual_results_generalisation}
\end{table*}

\begin{table*}
\centering
\resizebox{0.85\linewidth}{!}{
    \noindent
    \setlength\tabcolsep{4pt}
    \renewcommand{\arraystretch}{1.1}
    \begin{tabular}{l |c |c || cc |cc |cc |cc}
    \hline
        \multirow{3}{*}{\textbf{System}} & \multirow{3}{*}{\textbf{Model}} & \multirow{3}{*}{\textbf{\makecell[bc]{Train size\\(FEVER)}}}& \multicolumn{2}{c|}{\textbf{DanFEVER}}&\multicolumn{2}{c|}{\textbf{CHEF}}&\multicolumn{2}{c|}{\textbf{Unified-FC}}&\multicolumn{2}{c}{\textbf{RU22Fact}}\\
 &  && \multicolumn{2}{c|}{Da}& \multicolumn{2}{c|}{Zh}& \multicolumn{2}{c|}{Ar}& \multicolumn{2}{c}{Ru/Ukr}\\
        &  && F1 &  Acc     &F1 &Acc    &F1 &Acc    &F1 &Acc \\
    \hline
    ProoFVer            & mBART&145K& 36.12&  55.22&20.18&37.72   &39.67&48.04 &51.77&81.68 \\
    QA-NatVer         & mT0 &64& \tbf{63.64} &  \tbf{68.41} &-&-   &-&- &-&- \\
    \tbf{\projectname{}}  & Llama3-8B    &None& 53.9& 62.55& \tbf{47.94}& \tbf{53.2}   &\tbf{57.23}&\tbf{57.35} &\tbf{79.89}&\tbf{86.57} \\
    \hline
    \end{tabular}
}
\caption{\textbf{Zero-shot transfer results across non-English multilingual datasets.} Macro-F1 and accuracy scores for systems trained on the FEVER dataset. For each system, we report the provided language model and the size of the training data. QA-NatVer currently does not support non-English languages, except for Danish.}
\label{tab:multilingual_results_transfer}
\end{table*}

\subsection{Multilingual Experiments}

Our experimental results on non-English datasets in the zero-shot generalization and transfer setups are presented in Tables~\ref{tab:multilingual_results_generalisation} and~\ref{tab:multilingual_results_transfer}, respectively.

\paragraph{Generalization}

As shown in the results, \projectname{} outperforms both NatLog-based baselines, ProoFVer and QA-NatVer, across all datasets in the generalization setup. \projectname{} also demonstrates competitive performance compared to Direct-QA. Since QA-NatVer uses separate, language-specific models for text chunking, our experiments were limited to the available chunkers, specifically for English and Danish. This limitation highlights \projectname{}'s broader applicability, as it leverages a single model without requiring additional components like a chunker. Lastly, \citet{dubey2024llama} note that while Llama3-8B was pre-trained on multilingual data, it was primarily intended for English. This may explain the weaker performance of some systems using the model. However, \projectname{} still achieved better results compared to other baselines with multilingual backbones like mBART and mT0.

\paragraph{Transfer}

In the transfer setup, \projectname{} outperforms ProoFVer with a multilingual backbone across all datasets but falls behind QA-NatVer on DanFEVER, where QA-NatVer achieves \num{5.86} more accuracy points. Nonetheless, our results show strong performance, especially given that the baselines use multilingual models and are directly trained on NatLog data.


\subsection{Further Experiments}

\paragraph{Ensemble Size}

\begin{figure}
\centering
\includegraphics[width=0.8\linewidth, trim={0.2cm 0.3cm 0.3cm 0.4cm},clip]{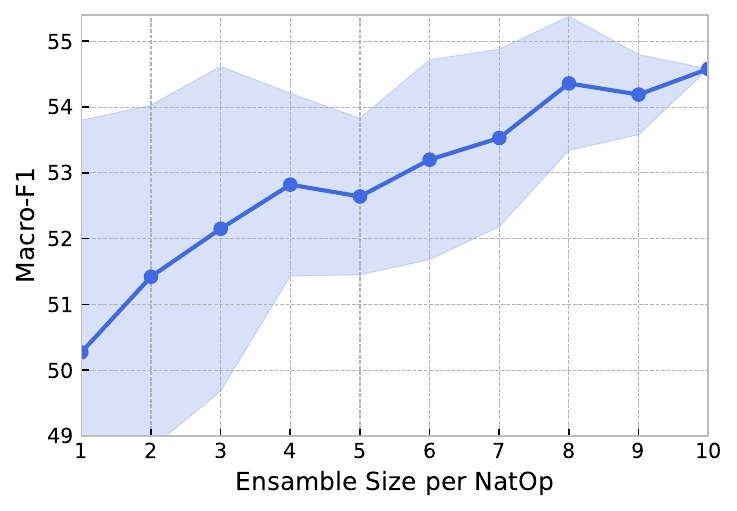}
\caption{The averaged Macro-F1 scores for different ensemble sizes, calculated from 20 independent runs for each size.}
\label{fig:ensemble_size}
\end{figure}

\begin{table*}[t]
\centering
\resizebox{0.85\linewidth}{!}{
    \setlength\tabcolsep{5pt}
    \renewcommand{\arraystretch}{1.1}
    \begin{tabular}{l |cc |cc |cc |cc}
    \hline
        \multirow{3}{*}{\textbf{System}} & \multicolumn{2}{c|}{\textbf{C-FEVER}} & \multicolumn{2}{c|}{\textbf{SciFact}}  & \multicolumn{2}{c|}{\textbf{PubHealth}}  & \multicolumn{2}{c}{\textbf{Hover}}\\
        &  F1 & Acc & F1 & Acc & F1 & Acc & F1 & Acc    \\
    \hline
    \tbf{\projectname{}}  & 46.02& 51.12& 54.58& 58.33& 69.21& 70.01& 60.26& 60.27\\
\hline
    - weighted templates  & 45.72& 50.40& 54.28 & 58.00   & 68.51& 69.30& 60.22 & 60.22 \\
    - QA templates  & 40.60& 49.89& 46.49& 52.00& 68.20& 69.20& 57.17& 57.50 \\
    - constrained decoding  & 41.85& 45.69& 52.65& 57.00& 65.26& 66.46& 59.26& 59.30 \\
    - alignment signals  & 40.62& 43.66& 52.27& 55.00& 54.94& 55.22& 58.72& 58.73 \\
\hline
    \end{tabular}
}
\caption{\textbf{Ablation study of \projectname{}.}}
\label{tab:main_ablation_study}
\end{table*}


To assess the impact of prompt ensemble size (Section~\ref{sec:method_ensembles}), we conducted an experiment measuring performance for various ensemble sizes. For each ensemble size $S$, we randomly sampled $S$ prompts for each NatOp from our prompt bank. This process was repeated \num{20} times, and we report the means and standard deviations for each ensemble size in Figure~\ref{fig:ensemble_size}.
The results show that prompt ensemble size substantially affects the variability of outcomes. When using only one prompt per NatOp and sampling different prompts, the Macro-F1 scores have a standard deviation of \num{3.53} points. However, an ensemble of just four prompts reduces this variation by more than half.
While performance mostly improves as the ensemble size increases, a few sample instances performed better than the 10-prompt average.
This suggests that the performance of \projectname{} could be further improved by selecting the best-performing combination of prompts on a validation set. However, we refrained from using a validation set due to our zero-shot setup.

\begin{table}[H]
\centering
\resizebox{0.85\linewidth}{!}{
\begin{tabular}{l | c c }
\hline
                  & Macro-F1 & Accuracy    \\
\hline
Llama2-7B         & 20.57    & 41.67       \\
Llama2-13B        & 30.96    & 42.16       \\
Llama2-70B        & 57.47    & 60.33       \\
Llama3-8B         & 54.58    & 58.33       \\
GPT-3.5-Turbo     & 49.21    & 53.00       \\
\hline
\end{tabular}
}
\caption{SciFact results for LLMs of various sizes.}
\label{tab:model_size}
\end{table}

\paragraph{Model Size}

Table~\ref{tab:model_size} compares the performance of our method across various sizes and versions of Llama models, demonstrating a substantial improvement as the model scales up. We also evaluated our method using the proprietary model ChatGPT-3.5 \citep{openai2023gpt}. Although ChatGPT-3.5 is purportedly larger than Llama3-8B, our method performed better with the Llama model. This discrepancy may be due to API limitations, which prevented the use of constrained decoding and weighted prompting. Details on prompting are provided in Appendix~\ref{sec:appendix-prompting}.

\paragraph{Ablation Study}
As reported in Table~\ref{tab:main_ablation_study}, we performed four ablation studies to assess the importance of individual components in \projectname{}. First, we evaluated the performance without using weighted ensemble prompts and observed a slight decline of \num{0.45} accuracy points on average. Second, we ablated our method by omitting prompt ensembles and using a single randomly sampled prompt instead. This resulted in a drop in performance by \num{2.79} accuracy points, which aligns with our previous findings regarding ensemble sizes. Third, we ablated \projectname{} by using unconstrained generation during decoding, leading to an average accuracy drop of \num{2.82} points. Lastly, we ablated our method by removing alignment signals, which caused a substantial drop of \num{6.78} accuracy points on average.

\section{Conclusion}

We have presented \projectname{}, a zero-shot fact verification method grounded in natural logic. Our method leverages the generalization capabilities of instruction-tuned LLMs and generates faithful justifications for proofs without relying on training data annotated with natural logic. We have evaluated \projectname{} in two zero-shot setups, outperforming our baselines on most datasets.
The ablation study shows the importance of individual design choices, and our comparison with the direct non-NatLog approach shows that natural logic provides competitive performance while providing explainability via faithful justifications. We hope that the methods and analyses presented here enable further progress toward improving the efficiency and explainability of fact verification systems.

\section*{Acknowledgement}
Rami Aly was supported by the Engineering and Physical Sciences Research Council Doctoral Training Partnership (EPSRC). Andreas Vlachos is supported by the ERC grant AVeriTeC (GA 865958).
\clearpage

\section*{Limitations}
\raggedbottom
Natural logic is useful for explainability but is less expressive than semantic parsing methods such as lambda calculus \citep{lambda-zettlemoyer-2005}. This paper doesn't address natural logic's limitations. Furthermore, our method generates proofs, which are meant to be processed by the DFA from left to right. Nevertheless, natural logic-based inference is not constrained to such execution.

\section*{Ethics Statement}

\paragraph{Intended Use and Misuse Potential.} Our models can potentially captivate a wider audience and substantially reduce the workload for human fact-checkers. Nevertheless, it is crucial to acknowledge the possibility of their exploitation by malicious actors. As such, we strongly advise researchers to approach them with caution.

\paragraph{Accuracy and Infallibility.} Our approach improves the clarity of FV models, enabling individuals to make better-informed decisions about trusting these models and their assessments. However, it is crucial for users to remain critical while interpreting the results of these systems and not mistake explainability for accuracy. We clarify that our evaluations do not determine the factual accuracy of a statement in the real world; instead, we use sources like Wikipedia as the basis for evidence. Wikipedia is a great collaborative resource, yet it has mistakes and noise of its own, similar to any encyclopedia or knowledge source. Therefore, we advise against using our verification system to make definitive judgments about the veracity of the assessed claims, meaning it should not be relied upon as an infallible source of truth.

\bibliography{anthology,custom}

\begin{thebibliography}{52}
\expandafter\ifx\csname natexlab\endcsname\relax\def\natexlab#1{#1}\fi

\bibitem[{AI@Meta(2024)}]{llama3modelcard}
AI@Meta. 2024.
\newblock \href {https://github.com/meta-llama/llama3/blob/main/MODEL_CARD.md} {Llama 3 model card}.

\bibitem[{Akhtar et~al.(2023)Akhtar, Schlichtkrull, Guo, Cocarascu, Simperl, and Vlachos}]{akhtar-etal-2023-multimodal}
Mubashara Akhtar, Michael Schlichtkrull, Zhijiang Guo, Oana Cocarascu, Elena Simperl, and Andreas Vlachos. 2023.
\newblock \href {https://doi.org/10.18653/v1/2023.findings-emnlp.361} {Multimodal automated fact-checking: A survey}.
\newblock In \emph{Findings of the Association for Computational Linguistics: EMNLP 2023}, pages 5430--5448, Singapore. Association for Computational Linguistics.

\bibitem[{Aly et~al.(2023)Aly, Strong, and Vlachos}]{aly2023qa}
Rami Aly, Marek Strong, and Andreas Vlachos. 2023.
\newblock Qa-natver: Question answering for natural logic-based fact verification.
\newblock \emph{arXiv preprint arXiv:2310.14198}.

\bibitem[{Angeli and Manning(2014)}]{angeli2014naturalli}
Gabor Angeli and Christopher~D Manning. 2014.
\newblock Naturalli: Natural logic inference for common sense reasoning.
\newblock In \emph{Proceedings of the 2014 conference on empirical methods in natural language processing (EMNLP)}, pages 534--545.

\bibitem[{Atanasova et~al.(2020)Atanasova, Simonsen, Lioma, and Augenstein}]{atanasova2020generating}
Pepa Atanasova, Jakob~Grue Simonsen, Christina Lioma, and Isabelle Augenstein. 2020.
\newblock Generating fact checking explanations.
\newblock \emph{arXiv preprint arXiv:2004.05773}.

\bibitem[{Baly et~al.(2018)Baly, Mohtarami, Glass, M{\`a}rquez, Moschitti, and Nakov}]{baly2018integrating}
Ramy Baly, Mitra Mohtarami, James Glass, Llu{\'\i}s M{\`a}rquez, Alessandro Moschitti, and Preslav Nakov. 2018.
\newblock Integrating stance detection and fact checking in a unified corpus.
\newblock \emph{arXiv preprint arXiv:1804.08012}.

\bibitem[{Brockman et~al.(2020)Brockman, Welinder, Murati, and OpenAI}]{openaiapi}
Greg Brockman, Peter Welinder, Mira Murati, and OpenAI. 2020.
\newblock Openai: Openai api.
\newblock \url{https://openai.com/blog/openai-api}.

\bibitem[{Chen et~al.(2022)Chen, Bao, Sun, Zhang, Chen, Zhou, Xiao, and Li}]{Chen_Bao_Sun_Zhang_Chen_Zhou_Xiao_Li_2022}
Jiangjie Chen, Qiaoben Bao, Changzhi Sun, Xinbo Zhang, Jiaze Chen, Hao Zhou, Yanghua Xiao, and Lei Li. 2022.
\newblock \href {https://doi.org/10.1609/aaai.v36i10.21291} {Loren: Logic-regularized reasoning for interpretable fact verification}.
\newblock \emph{Proceedings of the AAAI Conference on Artificial Intelligence}, 36(10):10482--10491.

\bibitem[{Chen et~al.(2023)Chen, Xu, Zeng, Sun, Li, and Xiao}]{chen2023converge}
Jiangjie Chen, Rui Xu, Wenxuan Zeng, Changzhi Sun, Lei Li, and Yanghua Xiao. 2023.
\newblock Converge to the truth: Factual error correction via iterative constrained editing.
\newblock In \emph{Proceedings of the AAAI Conference on Artificial Intelligence}, volume~37, pages 12616--12625.

\bibitem[{Chung et~al.(2022)Chung, Hou, Longpre, Zoph, Tay, Fedus, Li, Wang, Dehghani, Brahma et~al.}]{chung2022scaling}
Hyung~Won Chung, Le~Hou, Shayne Longpre, Barret Zoph, Yi~Tay, William Fedus, Yunxuan Li, Xuezhi Wang, Mostafa Dehghani, Siddhartha Brahma, et~al. 2022.
\newblock Scaling instruction-finetuned language models.
\newblock \emph{arXiv preprint arXiv:2210.11416}.

\bibitem[{De~Cao et~al.(2020)De~Cao, Izacard, Riedel, and Petroni}]{de2020autoregressive}
Nicola De~Cao, Gautier Izacard, Sebastian Riedel, and Fabio Petroni. 2020.
\newblock Autoregressive entity retrieval.
\newblock \emph{arXiv preprint arXiv:2010.00904}.

\bibitem[{Diggelmann et~al.(2020)Diggelmann, Boyd-Graber, Bulian, Ciaramita, and Leippold}]{diggelmann2020climate}
Thomas Diggelmann, Jordan Boyd-Graber, Jannis Bulian, Massimiliano Ciaramita, and Markus Leippold. 2020.
\newblock Climate-fever: A dataset for verification of real-world climate claims.
\newblock \emph{arXiv preprint arXiv:2012.00614}.

\bibitem[{Dubey et~al.(2024)Dubey, Jauhri, Pandey, Kadian, Al-Dahle, Letman, Mathur, Schelten, Yang, Fan et~al.}]{dubey2024llama}
Abhimanyu Dubey, Abhinav Jauhri, Abhinav Pandey, Abhishek Kadian, Ahmad Al-Dahle, Aiesha Letman, Akhil Mathur, Alan Schelten, Amy Yang, Angela Fan, et~al. 2024.
\newblock The llama 3 herd of models.
\newblock \emph{arXiv preprint arXiv:2407.21783}.

\bibitem[{Frantar et~al.(2022)Frantar, Ashkboos, Hoefler, and Alistarh}]{frantar2022gptq}
Elias Frantar, Saleh Ashkboos, Torsten Hoefler, and Dan Alistarh. 2022.
\newblock Gptq: Accurate post-training quantization for generative pre-trained transformers.
\newblock \emph{arXiv preprint arXiv:2210.17323}.

\bibitem[{Ganitkevitch et~al.(2013)Ganitkevitch, Van~Durme, and Callison-Burch}]{ganitkevitch2013ppdb}
Juri Ganitkevitch, Benjamin Van~Durme, and Chris Callison-Burch. 2013.
\newblock Ppdb: The paraphrase database.
\newblock In \emph{Proceedings of the 2013 Conference of the North American Chapter of the Association for Computational Linguistics: Human Language Technologies}, pages 758--764.

\bibitem[{Guo et~al.(2022)Guo, Schlichtkrull, and Vlachos}]{guo2022survey}
Zhijiang Guo, Michael Schlichtkrull, and Andreas Vlachos. 2022.
\newblock A survey on automated fact-checking.
\newblock \emph{Transactions of the Association for Computational Linguistics}, 10:178--206.

\bibitem[{Holtzman et~al.(2019)Holtzman, Buys, Du, Forbes, and Choi}]{holtzman2019curious}
Ari Holtzman, Jan Buys, Li~Du, Maxwell Forbes, and Yejin Choi. 2019.
\newblock The curious case of neural text degeneration.
\newblock \emph{arXiv preprint arXiv:1904.09751}.

\bibitem[{Hu et~al.(2022)Hu, Guo, Wu, Liu, Wen, and Yu}]{hu2022chef}
Xuming Hu, Zhijiang Guo, Guanyu Wu, Aiwei Liu, Lijie Wen, and Philip~S Yu. 2022.
\newblock Chef: A pilot chinese dataset for evidence-based fact-checking.
\newblock \emph{arXiv preprint arXiv:2206.11863}.

\bibitem[{Icard(2012)}]{icard2012inclusion}
Thomas~F Icard. 2012.
\newblock Inclusion and exclusion in natural language.
\newblock \emph{Studia Logica}, 100:705--725.

\bibitem[{Jacovi and Goldberg(2020)}]{jacovi2020towards}
Alon Jacovi and Yoav Goldberg. 2020.
\newblock Towards faithfully interpretable nlp systems: How should we define and evaluate faithfulness?
\newblock \emph{arXiv preprint arXiv:2004.03685}.

\bibitem[{Jiang et~al.(2023)Jiang, Ruan, Huang, Liao, Pitis, Grosse, and Ba}]{jiang2023calibrating}
Mingjian Jiang, Yangjun Ruan, Sicong Huang, Saifei Liao, Silviu Pitis, Roger~Baker Grosse, and Jimmy Ba. 2023.
\newblock Calibrating language models via augmented prompt ensembles.

\bibitem[{Jiang et~al.(2020)Jiang, Bordia, Zhong, Dognin, Singh, and Bansal}]{jiang2020hover}
Yichen Jiang, Shikha Bordia, Zheng Zhong, Charles Dognin, Maneesh Singh, and Mohit Bansal. 2020.
\newblock Hover: A dataset for many-hop fact extraction and claim verification.
\newblock \emph{arXiv preprint arXiv:2011.03088}.

\bibitem[{Kadavath et~al.(2022)Kadavath, Conerly, Askell, Henighan, Drain, Perez, Schiefer, Hatfield-Dodds, DasSarma, Tran-Johnson et~al.}]{kadavath2022language}
Saurav Kadavath, Tom Conerly, Amanda Askell, Tom Henighan, Dawn Drain, Ethan Perez, Nicholas Schiefer, Zac Hatfield-Dodds, Nova DasSarma, Eli Tran-Johnson, et~al. 2022.
\newblock Language models (mostly) know what they know.
\newblock \emph{arXiv preprint arXiv:2207.05221}.

\bibitem[{Kotonya and Toni(2020)}]{kotonya2020explainable_a}
Neema Kotonya and Francesca Toni. 2020.
\newblock Explainable automated fact-checking for public health claims.
\newblock \emph{arXiv preprint arXiv:2010.09926}.

\bibitem[{Krishna et~al.(2022)Krishna, Riedel, and Vlachos}]{krishna2022proofver}
Amrith Krishna, Sebastian Riedel, and Andreas Vlachos. 2022.
\newblock Proofver: Natural logic theorem proving for fact verification.
\newblock \emph{Transactions of the Association for Computational Linguistics}, 10:1013--1030.

\bibitem[{Lewis et~al.(2019)Lewis, Liu, Goyal, Ghazvininejad, Mohamed, Levy, Stoyanov, and Zettlemoyer}]{lewis2019bart}
Mike Lewis, Yinhan Liu, Naman Goyal, Marjan Ghazvininejad, Abdelrahman Mohamed, Omer Levy, Ves Stoyanov, and Luke Zettlemoyer. 2019.
\newblock Bart: Denoising sequence-to-sequence pre-training for natural language generation, translation, and comprehension.
\newblock \emph{arXiv preprint arXiv:1910.13461}.

\bibitem[{Li et~al.(2023)Li, Zhao, Chia, Ding, Bing, Joty, and Poria}]{li2023chain}
Xingxuan Li, Ruochen Zhao, Yew~Ken Chia, Bosheng Ding, Lidong Bing, Shafiq Joty, and Soujanya Poria. 2023.
\newblock Chain of knowledge: A framework for grounding large language models with structured knowledge bases.
\newblock \emph{arXiv preprint arXiv:2305.13269}.

\bibitem[{Lin et~al.(2022)Lin, Tan, Miller, Tian, and Ren}]{lin2022unsupervised}
Bill~Yuchen Lin, Kangmin Tan, Chris Miller, Beiwen Tian, and Xiang Ren. 2022.
\newblock Unsupervised cross-task generalization via retrieval augmentation.
\newblock \emph{Advances in Neural Information Processing Systems}, 35:22003--22017.

\bibitem[{Liu et~al.(2020)Liu, Gu, Goyal, Li, Edunov, Ghazvininejad, Lewis, and Zettlemoyer}]{liu2020multilingual}
Yinhan Liu, Jiatao Gu, Naman Goyal, Xian Li, Sergey Edunov, Marjan Ghazvininejad, Mike Lewis, and Luke Zettlemoyer. 2020.
\newblock Multilingual denoising pre-training for neural machine translation.
\newblock \emph{Transactions of the Association for Computational Linguistics}, 8:726--742.

\bibitem[{Muennighoff et~al.(2022)Muennighoff, Wang, Sutawika, Roberts, Biderman, Scao, Bari, Shen, Yong, Schoelkopf et~al.}]{muennighoff2022crosslingual}
Niklas Muennighoff, Thomas Wang, Lintang Sutawika, Adam Roberts, Stella Biderman, Teven~Le Scao, M~Saiful Bari, Sheng Shen, Zheng-Xin Yong, Hailey Schoelkopf, et~al. 2022.
\newblock Crosslingual generalization through multitask finetuning.
\newblock \emph{arXiv preprint arXiv:2211.01786}.

\bibitem[{Nakov et~al.(2021)Nakov, Corney, Hasanain, Alam, Elsayed, Barr{\'o}n-Cede{\~n}o, Papotti, Shaar, and Martino}]{nakov2021automated}
Preslav Nakov, David Corney, Maram Hasanain, Firoj Alam, Tamer Elsayed, Alberto Barr{\'o}n-Cede{\~n}o, Paolo Papotti, Shaden Shaar, and Giovanni Da~San Martino. 2021.
\newblock Automated fact-checking for assisting human fact-checkers.
\newblock \emph{arXiv preprint arXiv:2103.07769}.

\bibitem[{N{\o}rregaard and Derczynski(2021)}]{norregaard2021danfever}
Jeppe N{\o}rregaard and Leon Derczynski. 2021.
\newblock Danfever: claim verification dataset for danish.
\newblock In \emph{Proceedings of the 23rd Nordic conference on computational linguistics (NoDaLiDa)}, pages 422--428.

\bibitem[{OpenAI(2023)}]{openai2023gpt}
R~OpenAI. 2023.
\newblock Gpt-4 technical report. arxiv 2303.08774.
\newblock \emph{View in Article}, 2.

\bibitem[{Pan et~al.(2021)Pan, Chen, Xiong, Kan, and Wang}]{pan-etal-2021-zero}
Liangming Pan, Wenhu Chen, Wenhan Xiong, Min-Yen Kan, and William~Yang Wang. 2021.
\newblock \href {https://doi.org/10.18653/v1/2021.acl-short.61} {Zero-shot fact verification by claim generation}.
\newblock In \emph{Proceedings of the 59th Annual Meeting of the Association for Computational Linguistics and the 11th International Joint Conference on Natural Language Processing (Volume 2: Short Papers)}, pages 476--483, Online. Association for Computational Linguistics.

\bibitem[{Pan et~al.(2023{\natexlab{a}})Pan, Wu, Lu, Luu, Wang, Kan, and Nakov}]{pan-etal-2023-fact}
Liangming Pan, Xiaobao Wu, Xinyuan Lu, Anh~Tuan Luu, William~Yang Wang, Min-Yen Kan, and Preslav Nakov. 2023{\natexlab{a}}.
\newblock \href {https://doi.org/10.18653/v1/2023.acl-long.386} {Fact-checking complex claims with program-guided reasoning}.
\newblock In \emph{Proceedings of the 61st Annual Meeting of the Association for Computational Linguistics (Volume 1: Long Papers)}, pages 6981--7004, Toronto, Canada. Association for Computational Linguistics.

\bibitem[{Pan et~al.(2023{\natexlab{b}})Pan, Zhang, and Kan}]{pan2023investigating}
Liangming Pan, Yunxiang Zhang, and Min-Yen Kan. 2023{\natexlab{b}}.
\newblock Investigating zero-and few-shot generalization in fact verification.
\newblock \emph{arXiv preprint arXiv:2309.09444}.

\bibitem[{Perez et~al.(2021)Perez, Kiela, and Cho}]{perez2021true}
Ethan Perez, Douwe Kiela, and Kyunghyun Cho. 2021.
\newblock True few-shot learning with language models.
\newblock \emph{Advances in neural information processing systems}, 34:11054--11070.

\bibitem[{Popat et~al.(2018)Popat, Mukherjee, Yates, and Weikum}]{popat2018declare}
Kashyap Popat, Subhabrata Mukherjee, Andrew Yates, and Gerhard Weikum. 2018.
\newblock Declare: Debunking fake news and false claims using evidence-aware deep learning.
\newblock \emph{arXiv preprint arXiv:1809.06416}.

\bibitem[{Robertson and Walker(1994)}]{robertson1994some}
Stephen~E Robertson and Steve Walker. 1994.
\newblock Some simple effective approximations to the 2-poisson model for probabilistic weighted retrieval.
\newblock In \emph{SIGIR’94: Proceedings of the Seventeenth Annual International ACM-SIGIR Conference on Research and Development in Information Retrieval, organised by Dublin City University}, pages 232--241. Springer.

\bibitem[{Sanchez(1991)}]{sanchez1991studies}
Victor Sanchez. 1991.
\newblock \emph{Studies on natural logic and categorial grammar}.
\newblock Ph.D. thesis, University of Amsterdam.

\bibitem[{Schuster et~al.(2019)Schuster, Shah, Yeo, Filizzola, Santus, and Barzilay}]{schuster2019towards}
Tal Schuster, Darsh~J Shah, Yun Jie~Serene Yeo, Daniel Filizzola, Enrico Santus, and Regina Barzilay. 2019.
\newblock Towards debiasing fact verification models.
\newblock \emph{arXiv preprint arXiv:1908.05267}.

\bibitem[{Shu et~al.(2019)Shu, Cui, Wang, Lee, and Liu}]{shu2019defend}
Kai Shu, Limeng Cui, Suhang Wang, Dongwon Lee, and Huan Liu. 2019.
\newblock defend: Explainable fake news detection.
\newblock In \emph{Proceedings of the 25th ACM SIGKDD international conference on knowledge discovery \& data mining}, pages 395--405.

\bibitem[{Stacey et~al.(2023)Stacey, Minervini, Dubossarsky, Camburu, and Rei}]{stacey2023logical}
Joe Stacey, Pasquale Minervini, Haim Dubossarsky, Oana-Maria Camburu, and Marek Rei. 2023.
\newblock Logical reasoning for natural language inference using generated facts as atoms.
\newblock \emph{arXiv preprint arXiv:2305.13214}.

\bibitem[{Stacey et~al.(2022)Stacey, Minervini, Dubossarsky, and Rei}]{stacey-etal-2022-logical}
Joe Stacey, Pasquale Minervini, Haim Dubossarsky, and Marek Rei. 2022.
\newblock \href {https://doi.org/10.18653/v1/2022.emnlp-main.251} {Logical reasoning with span-level predictions for interpretable and robust {NLI} models}.
\newblock In \emph{Proceedings of the 2022 Conference on Empirical Methods in Natural Language Processing}, pages 3809--3823, Abu Dhabi, United Arab Emirates. Association for Computational Linguistics.

\bibitem[{Touvron et~al.(2023)Touvron, Martin, Stone, Albert, Almahairi, Babaei, Bashlykov, Batra, Bhargava, Bhosale et~al.}]{touvron2023llama}
Hugo Touvron, Louis Martin, Kevin Stone, Peter Albert, Amjad Almahairi, Yasmine Babaei, Nikolay Bashlykov, Soumya Batra, Prajjwal Bhargava, Shruti Bhosale, et~al. 2023.
\newblock Llama 2: Open foundation and fine-tuned chat models.
\newblock \emph{arXiv preprint arXiv:2307.09288}.

\bibitem[{Van~Benthem(1986)}]{VanBenthem1986}
Johan Van~Benthem. 1986.
\newblock \href {https://doi.org/10.1007/978-94-009-4540-1_6} {\emph{Natural Logic}}, pages 109--119. Springer Netherlands, Dordrecht.

\bibitem[{Vrande{\v{c}}i{\'c} and Kr{\"o}tzsch(2014)}]{vrandevcic2014wikidata}
Denny Vrande{\v{c}}i{\'c} and Markus Kr{\"o}tzsch. 2014.
\newblock Wikidata: a free collaborative knowledgebase.
\newblock \emph{Communications of the ACM}, 57(10):78--85.

\bibitem[{Wadden et~al.(2020)Wadden, Lin, Lo, Wang, van Zuylen, Cohan, and Hajishirzi}]{wadden2020fact}
David Wadden, Shanchuan Lin, Kyle Lo, Lucy~Lu Wang, Madeleine van Zuylen, Arman Cohan, and Hannaneh Hajishirzi. 2020.
\newblock Fact or fiction: Verifying scientific claims.
\newblock \emph{arXiv preprint arXiv:2004.14974}.

\bibitem[{Wright et~al.(2022)Wright, Wadden, Lo, Kuehl, Cohan, Augenstein, and Wang}]{wright-etal-2022-generating}
Dustin Wright, David Wadden, Kyle Lo, Bailey Kuehl, Arman Cohan, Isabelle Augenstein, and Lucy~Lu Wang. 2022.
\newblock \href {https://doi.org/10.18653/v1/2022.acl-long.175} {Generating scientific claims for zero-shot scientific fact checking}.
\newblock In \emph{Proceedings of the 60th Annual Meeting of the Association for Computational Linguistics (Volume 1: Long Papers)}, pages 2448--2460, Dublin, Ireland. Association for Computational Linguistics.

\bibitem[{Yao et~al.(2023)Yao, Zhao, Yu, Du, Shafran, Narasimhan, and Cao}]{yao2023react}
Shunyu Yao, Jeffrey Zhao, Dian Yu, Nan Du, Izhak Shafran, Karthik~R Narasimhan, and Yuan Cao. 2023.
\newblock \href {https://openreview.net/forum?id=WE_vluYUL-X} {React: Synergizing reasoning and acting in language models}.
\newblock In \emph{The Eleventh International Conference on Learning Representations}.

\bibitem[{Zeng et~al.(2024)Zeng, Ding, Zhao, Li, Zhang, Yao, Liu, and Qin}]{zeng2024ru22fact}
Yirong Zeng, Xiao Ding, Yi~Zhao, Xiangyu Li, Jie Zhang, Chao Yao, Ting Liu, and Bing Qin. 2024.
\newblock Ru22fact: Optimizing evidence for multilingual explainable fact-checking on russia-ukraine conflict.
\newblock \emph{arXiv preprint arXiv:2403.16662}.

\bibitem[{Zettlemoyer and Collins(2005)}]{lambda-zettlemoyer-2005}
Luke~S. Zettlemoyer and Michael Collins. 2005.
\newblock Learning to map sentences to logical form: Structured classification with probabilistic categorial grammars.
\newblock In \emph{Proceedings of the Twenty-First Conference on Uncertainty in Artificial Intelligence}, UAI'05, page 658–666, Arlington, Virginia, USA. AUAI Press.

\end{thebibliography}

\appendix

\begin{figure*}[!ht]
\noindent
\begin{lstlisting}[caption={Prompt template for the alignment task. Placeholders \textit{\{E\}} and \textit{\{C\}} get replaced by corresponding evidence and claim texts, respectively. Placeholders \textit{\{CH-1\}} to \textit{\{CH-N\}} get replaced by corresponding claim chunks, which were generated in the previous chunking step.}, label={prompt_chunk_and_align}]
CLAIM: {C}
EVIDENCE: {E}

-----
Align the following claim expressions with relevant substrings from the evidence text:
* {CH-1}
* {CH-2}
...
* {CH-N}

The aligned substrings should either support the expression, refute it, or simply refer to the same entity.
Where possible, provide an explanation following each alignment.
If no relevant alignment exists, write "None".
\end{lstlisting}
\end{figure*}


\section{Dataset Processing}
\label{sec:appendix_data}

To effectively assess the zero-shot capabilities of FV systems, it is important to evaluate the performance on real-life claims and consider domains requiring various domain expertise. We evaluated all models on datasets covering natural claims and domains such as climate change, biomedical subjects, government healthcare policies, and scientific literature. We chose datasets that mainly focus on three-way classification, i.e., using three labels \textit{Suppports}, \textit{Refutes}, or \textit{Not Enough Information}:



\paragraph{Climate-FEVER} \cite{diggelmann2020climate} dataset comprises \num{1535} real-life climate change claims, each annotated with five evidence sentences retrieved from Wikipedia. Each evidence sentence was labeled by five human annotators as supporting, refuting, or inconclusive regarding the claim's veracity, resulting in \num{5} votes for each evidence sentence. These votes were then aggregated to micro-verdicts for each retrieved evidence sentence, and micro-verdicts were further aggregated to a single macro-label for the claim. In our data processing, we combined all evidence sentences into a single paragraph and paired them with the macro-label assessment. Besides the standard three labels, some claims in the datasets are labeled as \textit{DISPUTED} if they are paired with both supporting and refuting micro-verdicts. Since our work focuses on three-label class prediction, we removed those \num{154} claims from the dataset.

\paragraph{PubHealth} \cite{kotonya2020explainable_a} is a dataset with natural claims in the public health domain. These claims are accompanied by evidence that requires subject matter expertise, along with expert explanations (judgments).
The dataset contains four labels \textit{True}, \textit{False}, \textit{Unproven}, and \textit{Mixture}. However, the classes are heavily unbalanced and the labels \textit{Unproven} and \textit{Mixture} cover less than 10\% of the data in total. Therefore, we use test set claims with only \textit{True} and \textit{False} labels, resulting in \num{987} claims paired with expert explanations as evidence.

\paragraph{SciFact} \cite{wadden2020fact} is a dataset of expert-written scientific claims paired with evidence that was extracted from academic papers. We collect the claims with supporting and refuting rationale and construct claim-evidence pairs with \textit{SUPPORT} and \textit{REFUTE} labels. Claims lacking a specific rationale are categorized as \textit{NEI}, and we pair them with the entire abstract text. We evaluate our pipeline on a test set that consists of \num{300} claims.

\paragraph{Hover} \cite{jiang2020hover} is an open-domain, multi-hop FV dataset, containing artificial claims built from the Wikipedia corpus. Its claims are labeled as either \textit{SUPPORTED} and \textit{NOT-SUPPORTED}. We use the development set, which consists of \num{4000} claims. In order to obtain evidence for all claims, we use the BM25 retriever \citep{robertson1994some}.

\paragraph{DanFEVER} \cite{norregaard2021danfever} is a Danish dataset of counterfactual claims constructed from Danish Wikipedia. It consists of \num{6407} instances and provides gold evidence for \textit{Supported} and \textit{Refuted} claims. To obtain evidence for \textit{NEI} claims, we use the BM25 retriever \citep{robertson1994some}.

\paragraph{CHEF} \cite{hu2022chef} is a Chinese dataset of real-world claims. We use their development set, which consists of \num{703} claims.

\paragraph{Unified-FC} \cite{baly2018integrating} is an Arabic dataset for fact-checking and stance detection. It contains \num{219} false claims from the VERIFY project\footnote{https://verify-sy.com/}, and \num{203} true claims from REUTERS\footnote{http://ara.reuters.com/}. Each claim in the dataset is paired with relevant articles retrieved via the Google Search API. For each claim, we concatenated all related articles and used them as gold evidence.

\paragraph{RU22Fact} \cite{zeng2024ru22fact} is a multilingual fact-checking dataset covering four languages: English, Chinese, Russian, and Ukrainian. For our multilingual study, we used their development set and extracted only claims in Russian and Ukrainian. While the original dataset classifies claims into three categories—\textit{Supported}, \textit{Refuted}, and \textit{Not Enough Information}—the Russian and Ukrainian claims were limited to just two labels: \textit{Supported} and \textit{Refuted}. As a result, our post-processed dataset consisted of \num{581} claims, and we approached the task as a binary classification problem.

\section{Baselines}
\label{sec:appendix_baselines}

\paragraph{ProoFVer} \cite{krishna2022proofver} is a seq2seq FV model that generates natural logic proofs as sequences of \textit{(claim, evidence, NatOp)} triples. ProoFVer is based on GENRE \citep{de2020autoregressive}, an end-to-end entity linking model that was obtained by fine-tuning the BART language model \citep{lewis2019bart}. ProoFVer was trained on a large collection of \num{145449} claims from FEVER that were heuristically annotated with natural logic proofs.
    
\paragraph{QA-NatVer} \cite{aly2023qa} is also based on natural logic but uses a question-answering framework to determine proofs. As a few-shot method, QA-NatVer was trained only on a small subset of FEVER data. It uses \num{64} training instances, which were further manually annotated with natural logic proofs.

QA-NatVer currently supports BART0 \citep{lin2022unsupervised}, Flan-T5 \citep{chung2022scaling} and mT0 \citep{muennighoff2022crosslingual} backbones.

\paragraph{Pan et al.} \citet{pan2023investigating} recently published an extensive analysis of zero-shot FV over 11 FV datasets. In their work, they experimented with different combinations of datasets for training and testing. While \citet{pan2023investigating} consider their experiments as zero-shot generalization tasks, in our work, we consider them as zero-shot transfer because they train their models on other FV datasets. Their results show useful zero-shot baselines over most of our datasets, providing a comparison with FV models that are not based on natural logic.

% ...

\section{Models}
\label{sec:appendix-models}



\paragraph{Llama models}
For experiments with Llama3 \citep{llama3modelcard}, we ran the \num{8}B parameter model in 16-bit precision for inference. For experiments with Llama2, we locally ran the \num{7}B, \num{13}B, and \num{70}B parameter models and used the GPTQ \citep{frantar2022gptq} version of these models with 4-bit quantization to reduce computational requirements and accelerate inference.

\paragraph{Hyperparameters}
When decoding with \mbox{Llama} models, we did not tune any hyper-parameters and used the values described in \citet{touvron2023llama}. Specifically, in the question-answering task for NatOPs, we set temperature to \num{1.0} and use nucleus sampling \cite{holtzman2019curious} with top-p set to \num{0.9}. For all other tasks, we change temperature to \num{0.1}.

\paragraph{Experimental Setup}
All experiments using Llama3 as the instruction-finetuned LLM were run on a machine with a single Quadro RTX 8000 with 49GB memory and 64GB RAM memory.



\section{Prompting}
\label{sec:appendix-prompting}

Listings\ref{prompt_chunk_and_align} show prompt templates for the evidence-rephrasing task, and the chunking and alignment task, respectively. These prompt templates were used for all experiments with Llama3 and ChatGPT models.

\paragraph{NatOp assignment}

Listing \ref{templates} shows the prompt templates used in the question-answering task for NatOps. Given a claim-evidence pair, we generated \num{10} distinct questions for each NatOp in separate prompts, replacing \textit{X} with the claim text and \textit{Y} with the evidence text. Additionally, we added the phrase \textit{"Answer Yes or No."} at the end of each prompt to encourage the \textit{Yes/No} output format. Lastly, we used the default system prompt \textit{"You are a helpful assistant."} for all prompts.

\paragraph{ChatGPT} We used OpenAI's API \citep{openaiapi} to query \textit{gpt-3.5-turbo-1106} and used the same hyperparamteres as with Llama3 models.
Due to the API limitations, we were unable to use constrained decoding for rephrasing, chunking, and alignment. Moreover, we could not use weighted prompt ensembles due to the inability to access the model's log-likelihood scores. Otherwise, we could replicate all the steps of our method with ChatGPT.

\begin{figure*}
\begin{lstlisting}[caption={Template questions for determining NatOps.}, label={templates}]
@Equivalence@
Is X a paraphrase of Y?
Are X and Y semantically equivalent in meaning?
Is the meaning of X effectively the same as Y?
Do X and Y function as synonyms or paraphrases of each other?
Does X serve as a paraphrase or an alternative expression for Y?
Are X and Y synonymous or nearly synonymous in meaning?
Do X and Y mean the same, without using external knowledge or assumptions?
Are X and Y semantically identical when considered independently of external knowledge?
Considering just X and Y, do these expressions have the same meaning?
Comparing X with Y, are they semantically equivalent based solely on their respective content?

@Entailment@
Given the premise Y does the hypothesis X hold?
Does the expression Y entail X?
Does the phrase Y logically imply X?
Is it true that if Y then X?
Is X a valid inference from Y?
Can X be inferred from the statement Y?
Given just the statements Y and X, does the first statement logically and necessarily imply the second without any external information?
Is it true that the statement Y logically entails X based solely on the information within the statements?
Does Y imply X when only the information within these statements is considered?
Is it accurate to say that Y categorically entails X, without external interpretations?

@Negation@
Is the phrase X a negation of Y?
Do X and Y represent mutually exclusive states, where the presence of one negates the possibility of the other?
Is the relationship between X and Y binary, such that if X is true, Y must necessarily be false, and vice versa?
Do X and Y negate each other completely?
Are X and Y in a dichotomous relationship, where the existence of one implies the non-existence of the other?
Is there a mutually exclusive relationship between X and Y, indicating that only one can be true at any given time?
In the context of X and Y, does the affirmation of one mean the automatic negation of the other?
Do X and Y form a binary opposition, where one categorically negates the other?
Are X and Y opposites in such a way that they cannot be true simultaneously?
Is the relationship between X and Y characterized by a strict either/or dichotomy?

@Alternation@
Does X exclude Y?
Do X and Y represent distinct alternatives, but not the only possibilities in their category?
Are X and Y exclusively different without negating the existence of additional states or options?
Do X and Y denote exclusive but not exhaustive options within a larger set of possibilities?
In comparing X and Y, are they distinct yet not limiting the possibility of other variations or alternatives?
Are X and Y distinct entities or states that exclude each other without forming a complete, exhaustive set?
Are X and Y different entities or states, but not in a way that negates the possibility of other, different entities or states?
Are X and Y distinct entities or states that exclude each other without forming a complete, exhaustive set?
In comparing X and Y, are they exclusive in nature but not necessarily covering all possible alternatives?
Do X and Y define separate, non-intersecting options, while not encompassing all possible scenarios?
\end{lstlisting}
\end{figure*}

\begin{figure*}
\noindent
\begin{lstlisting}[caption={Prompt template for FV experiments in a direct 
 multiple-choice setup. Placeholders \textit{\{E\}} and \textit{\{C\}} get replaced by corresponding texts.}, label={prompt_direct_mc}]
Given the claim "{C}" and the evidence "{E}", determine if the evidence supports, contradicts, or is insufficient to conclude about the claim.

Choices: 
(A): Supports
(B): Refutes
(C): Not Enough Information
\end{lstlisting}
\end{figure*}


\end{document}